\documentclass{article}

\pdfoutput=1
\usepackage{ijcai22}

\usepackage{times}

\usepackage{soul}
\usepackage{url}
\usepackage[hidelinks]{hyperref}
\usepackage[utf8]{inputenc}
\usepackage[small]{caption}
\usepackage{graphicx}
\usepackage{amsmath}
\usepackage{booktabs}
\usepackage{mathtools}
\usepackage{todonotes}
\usepackage{amsfonts}
\usepackage{paralist}
\usepackage{multicol}
\usepackage{xcolor}
\usepackage{subfig}
\usepackage{caption}
\definecolor{simplecolor}{rgb}{0.8, 0.0, 0.0}
\definecolor{envcolor}{rgb}{0.0, 0.0, 0.8}
\definecolor{comb1color}{rgb}{0.0, 0.4, 0.13}
\definecolor{comb2color}{rgb}{0.0, 0.0, 0.0}

\definecolor{twomovescolor}{rgb}{0.8, 0.0, 0.0}
\definecolor{complexcolor}{rgb}{0.0, 0.0, 0.8}
\definecolor{basiccolor}{rgb}{0.0, 0.4, 0.13}
\definecolor{fuzzcolor}{rgb}{0.0, 0.0, 0.0}

\usepackage{tikz}
\usepackage{pgfplots}
\pgfplotsset{compat=newest}
\usepackage[ruled,noend,linesnumbered]{algorithm2e}
\newcommand{\algorithmfontsize}{\scriptsize}

\SetCommentSty{mycommfont}
\newcommand{\RQone}{(Q1) \emph{Does the trained deep RL agent circumvent safety violations?}}
\newcommand{\RQtwo}{(Q2) \emph{Does the trained deep RL agent perform well from a wide variety of states? }}

\usepackage{graphicx}

\newcommand{\mdp}{\mathcal{M}}
\newcommand{\MdpTuple}[1][]{\ensuremath{(\states{#1},\sinit{#1},\Act{#1},\pmdp{#1},\rewFunction{#1})}}

\newcommand{\MdpInit}[1][]{\ensuremath{\mdp{#1}=\MdpTuple[#1]}}

\newcommand{\sinit}{s_0} 
\newcommand{\states}[1][]{\mathcal{S}_{#1}}
\newcommand{\act}[1][a]{\alpha} 
\newcommand{\Act}{\mathcal{A}}
\newcommand{\pmdp}{\mathcal{P}}
\newcommand{\rewFunction}{\mathcal{{R}}}

\newcommand{\trace}{\tau}
\newcommand{\traces}{\mathcal{T}}

\newcommand{\R}{\mathbb{R}}

\newcommand{\e}{\ensuremath{\mathbb{E}}}

\newcommand{\exec}{\mathit{exec}}

\newcommand{\response}[1]{\textcolor{black}{#1}}

\title{Search-Based Testing of Reinforcement Learning
\footnote{To appear in Proceedings of IJCAI-ECAI 2022,
$\:$ (\url{ijcai.org}).}}

\author{
Martin Tappler$^{1,3}$
\footnote{Contact author.}
\and Filip Cano C\'ordoba$^{2\:\dagger}$\and
Bernhard K.\ Aichernig$^{1,3}$\and
Bettina K\"onighofer$^{2,4\:\dagger}$\\
\affiliations
$^1$Institute of Software Technology, Graz University of Technology \\
$^2$Institute of Applied Information Processing and Communications, Graz University of Technology\\
$^3$TU Graz-SAL DES Lab,Silicon Austria Labs, Graz, Austria%
\\%
$^4$Lamarr Security Research%
\\%
\emails%
martin.tappler@ist.tugraz.at, %
filip.cano@iaik.tugraz.at, %
aichernig@ist.tugraz.at, %
bettina.koenighofer@lamarr.at %
\\ \vspace{0.5cm}
\includegraphics[width=17.5cm, height=1.4cm]{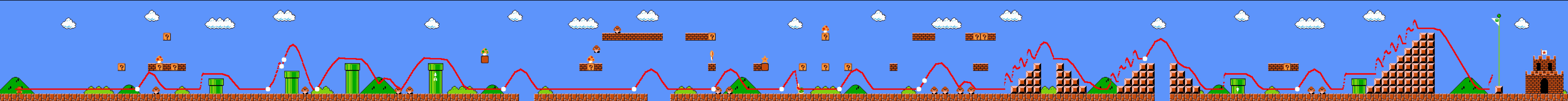}
\includegraphics[width=17.5cm, height=1.4cm]{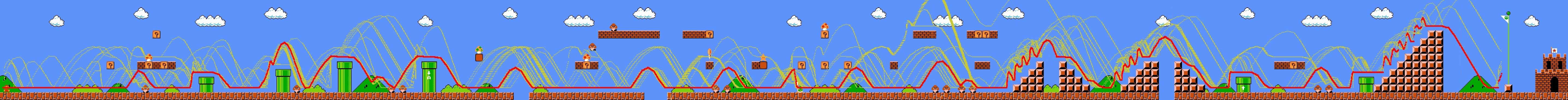}
\captionof{figure}{Super Mario Bros. Up: Reference Trace and Boundary States. Down: Reference Trace and Fuzz Traces.}
\addtocounter{figure}{-1}
\label{fig:mario}
}

\begin{document}

\maketitle


\begin{abstract}
Evaluation of deep reinforcement learning (RL) is inherently challenging. Especially the opaqueness of learned policies and the stochastic nature of both agents and environments make testing the behavior of deep RL agents difficult. We present a search-based testing framework that enables a wide range of novel analysis capabilities for evaluating the safety and performance of deep RL agents. For \emph{safety testing}, our framework utilizes a search algorithm that searches for a reference trace that solves the RL task. The backtracking states of the search, called \emph{boundary states}, pose safety-critical situations. We create safety test-suites that evaluate how well the RL agent escapes safety-critical situations near these boundary states. For \emph{robust performance testing}, we create a diverse set of traces via fuzz testing. These fuzz traces are used to bring the agent into a wide variety of potentially unknown states from which the average performance of the agent is compared to the average performance of the fuzz traces. We apply our search-based testing approach on RL for Nintendo's Super Mario Bros.\end{abstract}

\section{Introduction}

In reinforcement learning (RL)~\cite{DBLP:books/lib/SuttonB98}, an agent aims 
to maximize the total amount of reward through trial-and-error via interactions with
an unknown environment. Recently, RL algorithms
achieved stunning results in playing video games and complex board games~\cite{DBLP:journals/corr/abs-1911-08265}.

To achieve a broad acceptance 
and enlarge the application areas
of learned controllers, there is the urgent need to reliably evaluate trained RL agents.
When evaluating trained agents, two fundamental questions
need to be answered:
 \RQone ~\RQtwo
Testing deep RL agents is notoriously difficult. 
The first challenge arises from the environment, which is
is often not fully known and has
an immense state space,
combined with the byzantine complexity of the agent's model, and the lack of determinism of both the agent and the environment.
Secondly, to evaluate the performance of a trained agent's policy, an estimation of the performance of the optimal policy is needed. 

To address these challenges, we transfer well-established search-based concepts from software testing into the RL-setting. \emph{Search algorithms} like backtracking-based depth-first search (DFS) are standard  to find valid and invalid program executions. 
\emph{Fuzz testing} refers to automated software 
testing techniques that generate interesting test cases with the goal to expose corner cases that have not been properly dealt with in the program under test. 
In this work, we propose a search-based testing framework to reliably evaluate trained RL agents to answer the questions Q1 and Q2.
Our testing framework comprises four steps:

\noindent \textbf{Step 1: Search for reference trace and
boundary states.}
In the first step, we use a DFS algorithm to search for a \emph{reference trace} that solves the RL task by sampling the black-box environment. 
This idea is motivated by the experience from AI 
competitions, 
like the Mario AI and the NetHack Challenges \cite{Karakovskiy2012,kuettler2020nethack}, 
where best performers are symbolic agents, 
providing a reference solution of the task 
faster than neural-based agents.
Furthermore, since the DFS algorithm backtracks
when reaching an unsafe state in the environment, 
the search reveals safety-critical situations that we call \emph{boundary states}.

\noindent \textbf{Step 2: Testing for safety.}
To answer Q1, our testing framework computes \emph{safety test-suites}
that bring the agent into safety-critical situations near the boundary states.
Based on the ability of the agent to succeed in
these safety-critical situations we can evaluate the safety of agents.
\response{A safe agent should not 
violate safety regardless of the situation it faces.}

\noindent \textbf{Step 3: Generation of fuzz traces.}
As a basis for performance testing, our testing framework  applies a \emph{search-based fuzzing
method} to compute a diverse set of traces from the reference trace (Step 1) aiming for traces that gain high rewards and cover large parts of the state space.

\noindent \textbf{Step 4: Testing for performance.}
To answer Q2, we  
 create \emph{performance test-suites} from the fuzz traces to bring the agent into a diverse set of states within the environment. As performance metric we propose to point-wise 
compare the averaged performance gained by executing the agent's policy with the averaged performance gained by executing the fuzz traces.

\response{Our approach is very general and can be adapted to 
several application areas. In settings where initial traces are given, for example, stemming from demonstrations by humans, such traces can be used as a basis for fuzzing. Our approach only needs to be able to sample the environment as an oracle. Even in the case of partial observability, our testing framework can be successfully applied. This is the case since we only need the information on whether a trace successfully completed the task to be learned,
partially completed the task, or whether it violated safety.
Exact state information is not required. 
Fuzzing has been applied to test complex software systems, like operating system kernels, 
communication protocols, and parsers for programming lanuages~\cite{DBLP:journals/tse/ManesHHCESW21}. Hence, it offers 
scalability to large environments.
}

In our case study, we apply our framework to test the safety and performance of a set of deep RL agents trained to play Super Mario Bros. Fig.~\ref{fig:mario}
shows the reference trace (red) and boundary states (white points) computed in Step 1 and the
fuzz traces (yellow) from Step 3, computed in our case study.
\response{Since we consider the environment (as well as the trained agent, where a learned policy may need to break ties) to be probabilistic, we execute every test case a number of times and present the averaged results.}

\noindent \textbf{Related Work.}
%
\response{While RL has proven successful in solving many complex tasks~\cite{silver2016mastering} and often outperforms
classical controllers~\cite{kiran2021deep}, safety concerns prevent learned controllers from being widely used in safety-critical applications.
The research on \emph{safe RL} targets to guarantee safety during the training and the execution phase of the RL agent~\cite{garcia2015comprehensive}.
Safe RL has attracted a lot of attention in the formal methods community, culminating in a growing body of work on
the verification of trained 
networks~\cite{DBLP:conf/atva/Ehlers17,Pathak2017VerificationAR,DBLP:conf/uai/CorsiMF21}.
However, all of these approaches suffer from scalability issues and are not yet able to verify industrial-size deep neural networks.
An alternative line of research aims to enforce safe operation of an RL agent during runtime, using techniques from runtime monitoring and enforcement ~\cite{DBLP:conf/aaai/AlshiekhBEKNT18,DBLP:conf/amcc/PrangerKTD0B21}.
These methods typically require a complete and faithful model of the environment dynamics, which is often not available.
While a large amount of work on offline and runtime verification of RL agents exists, studying suitable testing methods for RL has attracted less attention.
}

The development of RL algorithms has greatly benefited from benchmark
environments for performance evaluation, including the Arcade Learning Environment~\cite{DBLP:journals/jair/BellemareNVB13}, and OpenAI Gym~\cite{brockman2016openai}, Deepmind Control Suite~\cite{tassa2018deepmind}, to name a few. SafetyGym~\cite{Achiam2019BenchmarkingSE} was especially designed to evaluate the safety of RL algorithms during exploration.
Most work on testing for RL evaluates the aggregate performance by comparing the mean and median scores across tasks.
Recently, testing metrics addressing the statistical uncertainty in such point estimates have been proposed~\cite{DBLP:journals/corr/abs-2108-13264}. 
We extend previous work by proposing search-based testing tailored toward (deep) RL. We use search-based methods to automatically create safety-critical test-cases and test cases for robust performance testing.

RL has been proposed for software testing and 
in particular also for fuzz testing~\cite{deep_reinf_fuzzing,DBLP:conf/uss/WangZZQKA21,DBLP:conf/fm/ScottSRMG21,DBLP:journals/corr/abs-1807-07490}.
In contrast, we propose a novel search-based 
testing framework including fuzzing to test RL agents. Fuzzing
has been applied to efficiently solve complex tasks
~\cite{DBLP:conf/sp/AschermannSAH20,DBLP:journals/corr/abs-2111-03013}.
We
perform a backtracking-based search to efficiently solve the task, while fuzzing 
serves to cover a large portion of the state space. 
Related is also the work from Trujillo et al.~\cite{DBLP:conf/icse/TrujilloLEDC20} which analyzes the adequacy of neuron coverage for testing deep RL, whereas
our adequacy criteria are inspired by traditional boundary value and combinatorial testing. 

\response{
We used our testing framework to evaluate trained deep-Q learning agents, 
agents that internally use deep neural networks to approximate the Q-function.
Recent years have seen a surge in works on testing deep neural networks. 
Techniques, like DeepTest~\cite{DBLP:conf/icse/TianPJR18}, DeepXplore~\cite{DBLP:journals/cacm/PeiCYJ19}, and DeepRoad~\cite{DBLP:conf/kbse/ZhangZZ0K18}, are orthogonal to our proposed 
framework. While we focus on the stateful reactive nature of RL agents, viewing 
them as a whole, these techniques are used to test sensor-related aspects of autonomous agents and find applications in particular
in image processing.
Furthermore, we may consider taking into account neural-network-specific testing
criteria~\cite{DBLP:conf/kbse/MaJZSXLCSLLZW18}. However, doubts about the 
adequacy of neuron coverage and related criteria have been raised recently~\cite{DBLP:conf/sigsoft/Harel-CanadaWGG20}. Hence, more research
is necessary in this area, as has been pointed out by Trujillo et al.~\cite{DBLP:conf/icse/TrujilloLEDC20}.}

\noindent \textbf{Outline.} The remainder of the paper is structured as follows.
In Sec.~\ref{sec:prelim}, we give the background and notation. In the Sec.~\ref{sec:searchAlgo} 
to~\ref{sec:performance} 
we present and discuss in detail Step 1 - Step 4 of our testing framework.
We present a detailed case study in Sec.~\ref{sec:experiments}.

\section{Preliminaries}
\label{sec:prelim}

A \textbf{Markov decision process} (MDP) $\MdpInit$ is a tuple
with a finite set $\states$ of states including initial state $s_0$, a finite set $\Act=\{a_1\dots, a_n\}$ of actions, and
a \emph{probabilistic transition function} $\pmdp: \states \times \Act \times \states
\rightarrow [0,1]$,
and an \emph{immediate reward function} $\rewFunction: \states \times \Act \times \states \rightarrow \R$. 
For all $s \in \states$ the available actions are $\Act(s) = \{a \in \Act\: |\: \exists s', \pmdp(s, a, s') \neq 0\}$ and we assume $|\Act(s)| \geq 1$. 
A memoryless deterministic policy $\pi: \states \rightarrow \Act$ is a function over action given states. The set of all memoryless
deterministic policies is denoted by $\Pi$.

An \textbf{MDP with terminal states} is an MDP $\mdp$ with a set of \emph{terminal states}
$\states[T] \subseteq \states$ in which the MDP terminates, i.e., 
the execution of a policy $\pi$ on $\mdp$ yields a trace 
$exec_{\pi}(\pi, s_0) = \langle s_0, a_1, r_1, s_1, \dots, r_n, s_n\rangle$ with only $s_n$ being a state in $\states[T]$. 
$\states[T]$ consists of two types of states:
\emph{goal-states} $\states[G] \subseteq \states[T]$ representing states in which the task to be learned was accomplished by reaching them, and undesired
\emph{unsafe-states} $\states[U] \subseteq \states[T]$.
\response{A \emph{safety violation} occurs whenever a state in 
$\states[U]$ is entered.}
%
We define the set of \emph{bad-states} $\states[B]$ as all states that almost-surely 
lead to an unsafe state in $\states[U]$, i.e., a state $s_B\in \states$ is in $\states[B]$,
if applying any policy $\pi \in \Pi$ starting in $s_B$ leads to a state in $\states[U]$ with probability $1$.
The set of \emph{boundary-states} $\states[BO]$ is defined as the set of not bad states with successor states within the bad states, i.e., a state $s_{BO}\in \states$ is in $\states[BO]$ if $s_{BO}\not\in\states[B]$
and there exists a state $s\in \states[B]$ and an action $a\in \Act$ with $\mathcal{P}(s_{BO},a,s)>0$.

We consider \textbf{reinforcement learning} (RL) 
in which an agent learns 
a task through trial-and-error via interactions with
an unknown environment modeled by a MDP $\MdpInit$ with terminal states $\states[T]\in\states$. 
At each step $t$, the agent receives an observation
$s_t$. It then chooses an action $a_{t+1} \in \Act$. The environment then
moves to a state $s_{t+1}$ with probability $\pmdp(s_t, a_{t+1}, s_{t+1})$.
The reward is determined with $r_{t+1} = \rewFunction(s_t, a_{t+1}, s_{t+1})$.
If the environment enters a terminal state in $\states[T]$, the training episode ends.
The time step the episode ends is denoted by $t_{\mathrm{end}}$. 
The return $\texttt{ret}=\Sigma^{t_{\mathrm{end}}}_{t=1} \gamma^t r_t$ is the cumulative future discounted reward per episode, using the \emph{discount factor}
$\gamma \in [0,1]$. 
The objective of the agent is to learn an
\emph{optimal policy} $\pi^* : \states \rightarrow \Act$ that maximizes the expectation of the
return, i.e., $\max_{\pi\in\Pi} \e_\pi(\texttt{ret})$.
The accumulated reward per episode is $R=\Sigma^{t_{\mathrm{end}}}_{t=1} r_t$.

\textbf{Traces.}
A \emph{trace} $\trace= \langle s_0, a_1, r_1, s_1, \dots, a_n, r_n, s_n\rangle$ is the state-action-reward sequence induced by a policy during an episode starting with the initial state $s_0$.  We denote a set of traces with $\traces$.
%
%
%
Given a trace $\trace= \langle s_0, a_1, r_1, s_1 \dots r_n, s_n\rangle$,
we use $\trace[i]$ to denote 
the $i$\textsuperscript{th} state of $\trace$ ($s_i=\trace[i]$), 
$\trace^{-i}$ to denote the \emph{prefix} of $\trace$ ($\trace^{-i}$ consists of all entries from $\trace$ from position $0$ to $i$) and we denote the trace $\trace^{+i}$ to be the \emph{suffix} of $\trace$ 
($\trace^{+i}$ consists of all entries from $\trace$ from position $i$ to $n$).
Given a trace $\trace= \langle s_0, a_1, r_1, s_1 \dots r_n, s_n\rangle$, we denote $|\trace|=n$ to be the length of the trace.
We denote the first appearance of state $s$ in
trace $\tau$ by $d(\trace,s)$ (if $d(\trace,s)=i$ then $\trace[i]=s$). 
%
%
We call the action sequence resulting from omitting the states and rewards from
$\trace$ an \emph{action trace} $\trace_\Act = \langle a_1, a_2, \dots a_{n}\rangle$. 
$\trace_{\Act}[i]$ gives the $i^{th}$ action, i.e., $a_i=\trace_{\Act}[i]$. 
Executing $\trace_\Act$ on $\mdp$ from $s_0$
yields a trace $\exec_{\tau}(\trace_\Act, s_0) = \langle s_0, a_1, r_1, s_1 \dots r_n, s_n\rangle$ with $n = |\trace_\Act|$.



\section{Step 1 - Search for Reference Trace and Boundary States}
\label{sec:searchAlgo}

The first step of our testing framework is to perform a \emph{search} for a reference trace $\trace_{\mathrm{ref}}$ that performs the tasks to be learned by the RL agent (not necessarily in the optimal way)
and to detect boundary-states $\states[B0]'\subseteq\states[B0]$ along the reference trace.

We propose to compute $\tau_{\mathrm{ref}}$ using a \emph{backtracking-based, depth-first search} (DFS) by sampling the MDP $\mdp$.
For the DFS, we abstract away stochastic behavior of $\mdp$ by exploring all possible 
behaviors in visited states by repeating actions sufficiently often~\cite{DBLP:conf/icgi/KhaliliT14}.
%
Assuming that $p=\mathcal{P}(s,a,s')$ is the smallest transition
probability greater $0$ for any $s,s'\in \states$ 
and $a \in \Act$ in $\mdp$, we 
compute the number of repetitions $rep$  required to match a confidence level $c$ via
$rep(c, p) = log(1 - c)/ log(1 - p)$. 
This ensures observing all possible states with
a probability of at least $c$.



%
\textbf{Example.} Assume that $p=0.1$ is the smallest probability $>0$ in $\mdp$.
To achieve a confidence level of $90\%$ that the search visited any reachable state, 
the DFS has to perform $rep(0.9, 0.1)= 22$ repetitions of any action in any state. 

\SetKwFor{RepTimes}{repeat}{times}{end}

\begin{algorithm}[t!]
\algorithmfontsize
    \caption{Search for Reference Trace $\tau_{\mathrm{ref}}$}\label{alg:search}
    \SetKwInOut{Input}{input}\SetKwInOut{Output}{output}\SetNoFillComment
    \Input{MDP $\MdpInit$, repetitions $rep$}
    \Output{$\tau_{\mathrm{ref}}, \:$ $\states[BO]'$ }

    $\mathcal{V_S} \leftarrow [s_0]; \:\:\:
    \mathcal{V_A} \leftarrow [\:]; \:\:\:
    Explored \leftarrow \emptyset;    \:\:\:
    success \gets \textit{false}$\;
        $\tau_{\mathrm{ref}} \leftarrow [s_0];\:\:\: 
    \states[BO]' \leftarrow \emptyset$\;
    
    \texttt{DFS}($s_0$)\;
    \If{$success$}{
    $s_{\mathrm{prev}} \leftarrow s_0$\;
    \For{$i \in 1,\dots, |\mathcal{V_A}|$}{
    $a, \: s \:\leftarrow\: \mathcal{V_A}[i],\: \mathcal{V_S}[i+1]$\;
    \If{$s\notin Explored$}{
    $r \leftarrow \mathcal{R}(s_{\mathrm{prev}},a, s)$\;
    \textbf{Push}$(\tau_{\mathrm{ref}}, \langle a,r,s\rangle)$\;
    $s_{\mathrm{prev}} \leftarrow s$\;
    \If{$\mathcal{V_S}[i+2] \in Explored$}{ 
    \tcc{\algorithmfontsize next state is a backtracking point} 
    $\states[BO]' \gets \states[BO]'\cup\{ s\}$\;
    }
    }
    }
    }

    \SetKwProg{Fn}{Function}{:}{}
    \SetKwFunction{DFS}{DFS}
    \Fn{\DFS{$s$}}{
    \If{$s\in\states[U]$}{
    $Explored \gets Explored \cup \{s\};\:\: \Return$\;
    }
    \If{$s\in\states[G]\;$ \textbf{or} $\; success$}{
     $success \gets true$;$\:\: \Return$\;}
    \For{$a \in \Act$}{
    \RepTimes{$rep$}{
    Sample $s'$ from $ \mathcal{P}(s,a)$\;
    \If{$s'\notin \mathcal{V_S}$}{
    \textbf{Push}($\mathcal{V_A}$, $a$) $;\:\:$
    \textbf{Push}($\mathcal{V_S}$, $s$)\;
    \texttt{DFS}($s'$)\;
    }
    }

    }
    \lIf{$\lnot success$}{
    $Explored \gets Explored \cup 
    \{s\}$}
    }
\end{algorithm}

\emph{Algorithm~\ref{alg:search}} gives the pseudo code of our search algorithm to compute $\tau_{\mathrm{ref}}\in\traces$ and a set of 
boundary-states $\states[B0]'\subseteq\states[B0]$. 
The list $\mathcal{V_{\states}}$ stores states that have already been visited, and $\mathcal{V_{\Act}}$ 
stores the executed actions leading to the corresponding states in $\mathcal{V_{\states}}$.
Every time the search visits an unsafe state
the algorithm backtracks. 
A non-terminal state $s$ is added to $Explored$ if the DFS backtracked to $s$ from all successor states. \response{By tracking visited states
in $\mathcal{V_S}$, 
we ensure that we do not explore a state twice along the same trace. That is, we use $\mathcal{V_S}$ to detect cycles.}
When visiting a goal state,
$\textsc{DFS}(s_0)$ terminates successfully.
In this case, $\tau_{\mathrm{ref}}$ is built from
the set of visited states that were not part of a backtracking branch of the search, i.e., $s\in \tau_{\states}$ if $s\in \mathcal{V_{\states}}$ and $s \not\in Explored$, with corresponding actions in $\mathcal{V_{\Act}}$. 
States $s\in \tau_{\mathrm{ref}}$ that have successor states $s'\in Explored$ are boundary states, i.e., $s\in\states[BO]'$.

\textbf{Example.}
Figure~\ref{fig:search_mdp} shows parts of an MDP $\mdp$
that was explored during a run of our search algorithm.
Found unsafe states are marked red.
After visiting $s_{10}\in \states[G]$ (green circle), the search function $\textsc{DFS}(s_0)$ returns with  $\mathcal{V_{\states}}\!=\! [s_0,\dots,s_{10}]$,
$\mathcal{V_{\Act}}\!=\![a,a,b,a,b,b,a,a,b,b]$
and $Explored=\{s_2,s_3,s_4,s_5,s_8,s_9\}$. 
The reference trace (omitting rewards) is
$\tau_{\mathrm{ref}}=\{s_0,a,s_1,b,s_6,a,s_7,b,s_{10}\}$ and
the subset of boundary states is
$\states[BO]'=\{s_1, s_7\}$ (blue circles).

\textbf{Optimizing Search.} 
Proper abstractions of the state space may be used to merge similar states, thereby pruning
the search space and enabling 
to find cycles in the abstract state space via the DFS.
Detecting cycles speeds up the search since the DFS backtracks when finding an edge to an already visited state.
An example for such an abstraction is omitting the execution time in the state space to merge states. 

\addtocounter{figure}{1}

\begin{figure}
\centering
\includegraphics[width=7.6cm]{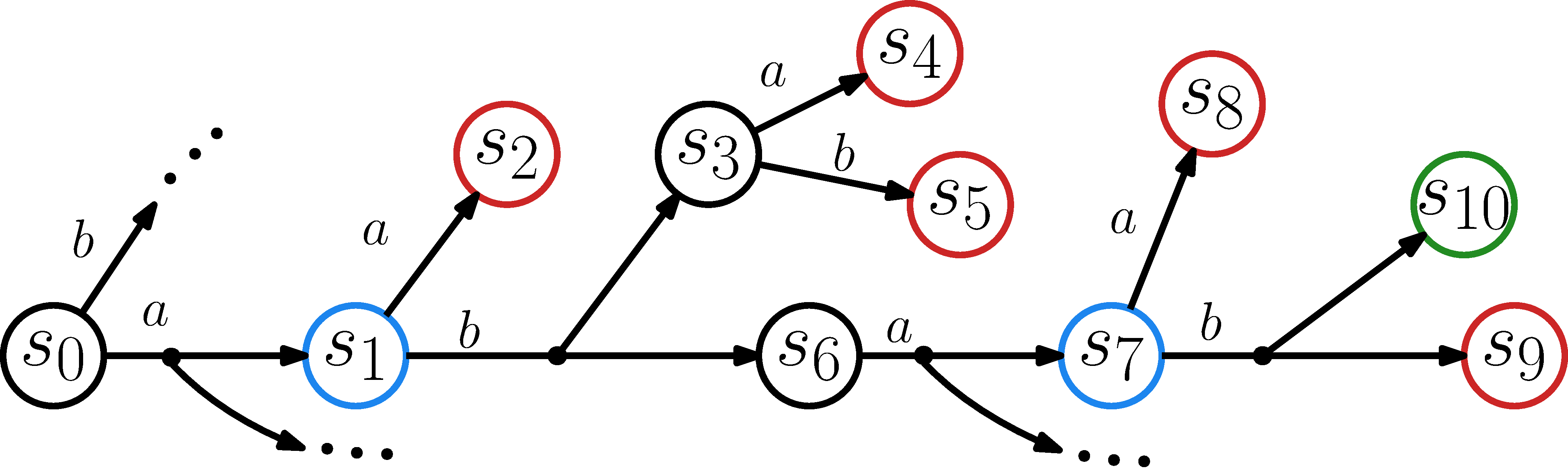}
    \caption{Run of the Search Algorithm}
    \label{fig:search_mdp}
\end{figure}



\section{Step 2 - Testing for Safety}

Based on $\tau_{\mathrm{ref}}$ and $\states[BO]'$ searched for in Step 1, we propose several \emph{test suites} to identify weak points of a policy with a high frequency of fail verdicts, i.e., safety violations.
After discussing suitable test suites, we discuss how to execute them to test the safety of RL agents.

\paragraph{Simple Boundary Test Suite.}
We use the boundary states  $\states[BO]'$ in $\tau_{\mathrm{ref}}$ for
boundary value testing~\cite{pezze_young_testing}.
We compute a simple test suite that consists of all prefixes of $\tau_{\mathrm{ref}}$ that end in a boundary state.
From these traces, we use the action traces  
to bring the RL agent into safety-critical situations and to test its behavior regarding safety.

Formally, let $DB$ be the sequence of depths of the boundary states $\states[BO]'$ in $\tau_{\mathrm{ref}}$, i.e., for any $s_{BO}\in \states[BO]' : d(\tau, s_{BO})\in DB$.
Using $DB$, we compute a set of traces $T$ by 
$$T=\{\tau_{\mathrm{ref}}^{-DB[i]} \mid 1 \leq i \leq |DB| \}.$$
Omitting states and rewards from the traces in $T$ results 
in a set of action traces that form a \emph{simple boundary test suite} called $ST$.
We say the action trace $\tau_{\Act,\mathrm{ref}}^{DB[i]}\in ST$
is the test case for the $i$\textsuperscript{th} boundary state in $\states[BO]'$.

\paragraph{Test Suites using Boundary Intervals.}
Boundary value testing checks not only boundary values,
but also inputs slightly off of the boundary~\cite{pezze_young_testing}.
To transfer this concept into RL testing, 
we introduce boundary intervals to test at additional states near the boundary.

In contrast to boundary testing of conventional software, our test cases stay in states traversed by $\tau_{\mathrm{ref}}$.
This choice is motivated by the definition of a boundary state: a state with successor states that necessarily lead to an unsafe state.
Bringing the RL agent in such a losing position will not provide additional insight concerning the learned safety objective, since the agent has no other choice than to violate safety. 
However, testing states of $\tau_{\mathrm{ref}}$ within an offset of boundary states provides insights into
how well the RL agent escapes safety-critical situations. 
Given a simple test suite $ST$ and an interval-size $is$, we create an 
\emph{interval test suite} $IT(is)$ by 
adding additional test cases to $ST$, such that 
$$IT(is) = \{\tau_{\Act,\mathrm{ref}}^{DB[i] + \mathit{off}} \:|\: \tau_{\Act,\mathrm{ref}}^{DB[i]} \in ST, 
-is \leq \mathit{off} \leq is\},$$ where $\tau_{\Act,\mathrm{ref}}$ is the reference action-trace.
The test case $\tau_{\Act,\mathrm{ref}}^{DB[i] + \mathit{off}}$ tests the agent at boundary
state $i$ with offset $\mathit{off}$.

\paragraph{Test Suites using Action Coverage.}
Combinatorial testing covers issues resulting from combinations
of input values. We adapt this concept by creating test suites that cover combinations of actions near boundary states, i.e., the test suite evaluates which actions cause unsafe behavior in boundary regions.
%
%
Given the reference action-trace $\tau_{\Act,\mathrm{ref}}$, a simple test suite $ST$, and a $k \geq 1$, we generate a  
\emph{$k$-wise action-coverage} test suite $AC(k)$ by creating $|\Act|^k$
test cases for every test case in $ST$ covering all $k$-wise combinations
of actions at the $k^{th}$ predecessor of a boundary state.
The test suite is given by $$AC(k) = 
\{\tau_{\Act,\mathrm{ref}}^{DB[i] -k} \cdot \mathit{ac} \:|\: \tau_{\Act,\mathrm{ref}}^{DB[i]} \in ST, 
\mathit{ac} \in \Act^k\}.$$


\paragraph{Test-case Execution \& Verdict.}

To test the behavior of an agent regarding safety, we use a safety test-suite to bring the agent in safety critical situations. 
A single test-case execution takes an action-trace $\trace_\Act$,
an inital state $s_0$ and a test length $l$ as parameters. 
To test an RL agent using $\trace_\Act$, we first execute $\trace_\Act$,
yielding a trace $exec(\trace_{\Act})=\langle s_0, a_1, r_1, s_1, \dots , a_n, r_n,s_n \rangle$.
A test case $\trace_\Act$ is \emph{invalid} if $exec(\trace_\Act)$ consistently visits a terminal state in $\states[T]$ when executed repeatedly. 

Starting from $s_n$, we pick the next $l$ actions according to the policy of the agent. 
Note that $l$ should be chosen large enough to evaluate the behavior of the agent regarding safety.
Therefore, it should be considerably larger than the shortest path to the next unsafe state in $S_U$.
%
After performing $l$ steps of the agent's policy, we evaluate the test case.
A test can \emph{fail} or \emph{pass}: A test fails, if starting from $s_n$ the agent reaches an unsafe state 
in $\states[U]$ within $l$ steps. Otherwise, the test passes. 


To execute a test suite $T$, we perform
every test case of $T$ 
$n$ times. 
During that, 
we compute 
the \emph{relative frequency of fail verdicts} resulting from 
executing each individual test case.

\section{Step 3 - Generation of Fuzz Traces}
Our testing framework evaluates the performance of RL agents
using fuzz traces. The traces are used to
compare gained rewards as well as to bring the agent in a variety of states and to evaluate the performance from these onward. 
In this section, we discuss 
the fuzz-trace generation for performance testing.
For this purpose, we propose a search-based 
fuzzing method~\cite{fuzzingbook2021} based on 
genetic algorithms. The goal 
is to find action traces that (1) cover a large portion
of the state space
while (2) accomplishing the task to be learned by the RL agent.

\noindent\textbf{Overview of Computation of Fuzz Traces.}
Given the reference trace $\tau_{\mathrm{ref}}$  that solves the RL task (i.e., $s_n \in \states[G]$) and 
parameter values for the number of generations $g$ and the population size $p$, 
the \emph{fuzz traces} are 
computed as follows:
\begin{compactenum}
    \item Initialize $\traces_0$, the trace population:
    $~\traces_0 \coloneqq  \{\tau_{\Act, \mathrm{ref}}\}$.
    \item For $i = 1\textbf{ to } g$ generations do:
    \begin{compactenum}
    \item Create $p$ action traces (called \emph{offspring}) from $\traces_{i-1}$ to yield a new population $\traces_i$ of size $p$ by:
        \begin{compactitem}
            \item \emph{either} \emph{mutating} a single parent trace from $\traces_{i-1}$,
            \item  \emph{or} through
        \emph{crossover} of two parents from $\traces_{i-1}$ with a specified crossover probability.
        \end{compactitem}
        \item Evaluate the \emph{fitness} of every offspring trace in $\traces_i$. 
    \end{compactenum}
    \item Return $\traces_\mathit{fit}$ containing the fittest trace of each generation
\end{compactenum}

The \emph{fitness} of a trace is defined in terms of
state-space coverage and 
the degree to which the RL task is solved.
The computation of the fuzz traces searches
iteratively for traces with a high fitness by choosing parent traces with a probability proportional 
to their fitness.
To promote diversity, we favor mutation over crossover, by setting
the crossover probability to a value $<0.5$.
The set of the fittest traces $\traces_\mathit{fit}$ will be used in 
Step 4 for performance testing. Using the single fittest trace
from every generation helps enforcing variety.

\noindent\textbf{Fitness Computation.}
We propose a fitness function especially suited for testing RL agents. For an action trace $\tau_{\Act}$,
the fitness $F(\tau_{\Act})$ is the weighted sum
of three normalized terms:
\begin{compactitem}
    \item The \emph{positive-reward term}  $r_{pos}(\tau_{\Act}, s_0)$
    is the normalized positive reward gained in $exec_\tau(\tau_{\Act}, s_0)$. 
    \item The \emph{negative-reward term} $r_{neg}(\tau_{\Act}, s_0)$ is the 
    normalized inverted negative reward gained in $exec_\tau(\tau_{\Act}, s_0)$.
    \item The \emph{coverage fitness-term} $fc(\tau_{\Act}, s_0)$
    describes the number of newly visited states by
    $exec_\tau(\tau_{\Act}, s_0)$, 
    normalized by dividing
    by the maximum number of newly visited states by any
    action traces in the current population.
\end{compactitem}

\emph{Positive rewards} correspond to the degree as to which the RL task is solved by $\tau_{\Act}$.
\emph{Negative rewards} often correspond to the time required to solve the RL task. Hence if $\tau_{\Act}$ solves the task fast, it would be assigned a small negative reward. We invert the negative reward 
to have only positive fitness terms. 
\response{We normalize $r_{pos}$/$r_{neg}$ by dividing it
by the highest $r_{pos}'$/$r_{neg}'$ in the current generation.} 
The \emph{coverage fitness-term} depends on all states visited in previous populations.
Assume that the current generation is $i$. 
Let $Cov_{ppop}$ be the set of all states visited by the previous populations $\bigcup_{j < i} \traces_j$ and let $Cov(\tau_{\Act})$ be the visited states when executing an action trace $\tau_{\Act}$, i.e.,
$Cov(\tau_{\Act}) = \bigcup_{k \leq n}\{s_k\}$,
where $exec_\tau(\tau_{\Act}, s_0) = \langle s_0, a_0, r_0, s_1, \dots, s_n \rangle$.

The \emph{coverage fitness-term} $fc(\tau_{\Act})$ is then given by 
$$fc(\tau_{\Act}) = \frac{|Cov_{ppop} \setminus Cov(\tau_{\Act})|}{\max_{\tau_{\Act}'\in \traces_i}{|Cov_{ppop} \setminus Cov(\tau_{\Act}')|}}.$$ 
$fc(\tau_{\Act})$ 
is a normalized value $\leq 1$ and changes during fuzzing as more states are covered.
The \emph{fitness} $F(\tau_{\Act})$ is given by
$$
F(\tau_{\Act}) = \lambda_{cov} fc(\tau_{\Act}) +
\lambda_{pos} r_{pos}(\tau_{\Act}) + 
\lambda_{neg} (1- |r_{neg}(\tau_{\Act})|), 
$$
where the factors $\lambda_j$ are weights and
$r_{pos}(\tau_{\Act})$ and $r_{neg}(\tau_{\Act})$ are
normalized rewards gained when executing $\tau_{\Act}$.

\textbf{Mutation \& Crossover.} 
To generate a new action trace,
we perform either a crossover of two parent traces or
mutate a single parent trace. 
For \emph{crossover}, we create a new offspring trace  splitting two parent traces and concatenating the resulting subtraces. Let $\tau_{\Act,1}$ and $\tau_{\Act,2}$ be the parent action-traces.
To create an offspring trace, we uniformly select a random \emph{crossover point} $i\in\{1,\dots ,\min(|\tau_{\Act,1}|, |\tau_{\Act,2}|)-1\}$. 
The offspring is the concatenation of $\tau^{-i}_{\Act,1}$ and $\tau^{+i}_{\Act,2}$. 
For \emph{mutation}, we repeatedly apply mutation operators.
Given a parent trace $\trace_{\Act}$ and a parameter $ms$
defining the potential effect size of a mutation, we create an offspring $\trace'_{\Act}$ as following:
\begin{compactenum}
    \item Uniformly sample $x\in\{1,\dots,ms\}$.
    \item Chose a mutation operator parametrized with $x$ and perform the mutation on $\trace_{\Act}$ to create an action trace $\trace_{\Act}'$.
    \item Stop with a probability $p_{mstop} \in  (0,1]$ and return 
    $\trace_{\Act}'$. Otherwise, set $\trace_{\Act}\leftarrow \trace_{\Act}'$ and continue with Step 1.
\end{compactenum}


The applied mutation operators are (1) Insert, (2) Remove
(3) Change, and (4) Append. Each performs its eponymous
operation on an action sequence of length $x$ at a randomly
chosen index in the parent trace, except for Append.

\begin{algorithm}[t!]
\algorithmfontsize
    \caption{Performance Testing with Fuzz Traces}\label{alg:simple_performance_testing}
    \SetKwInOut{Input}{input}\SetKwInOut{Output}{output}\SetNoFillComment
    \Input{$\MdpInit$, policy $\pi$, fuzz traces $\traces_\mathit{fit}$,
 \# episodes $n_{ep}$}
    \Output{Avg. accumulated rewards of the agent $R_a$ and the fuzz traces $R_t$}
    \Return $R_t \gets $ \texttt{EvalTraces}($\traces_\mathit{fit}$, $s_0$, $n_{ep}$), 
    $R_a \gets $ \texttt{EvalAgent}($\pi$, $s_0$, $n_{ep}$)\;
    
    \SetKwProg{Fn}{Function}{:}{}
    \SetKwFunction{EvalTraces}{EvalTraces}
    \Fn{\EvalTraces{$\traces_\mathit{fit}$, $s_0$, $n_{ep}$}}{
    
    \For{$\tau_{\Act}\in \traces_\mathit{fit}$}{
    \For{$i\gets1$ \text{\normalfont\bf\ to } $n_{ep}$}{
    $\tau_i \gets exec_{\trace}(\tau_{\Act}, s_0) = \langle s_0, a_1, r_1 \dots s_n\rangle$\;
    $R_{t,\tau_{\Act},i}\gets\Sigma^{n}_{k=1} r_k$ with $r_k \in \tau_i$
    }
    }
    \Return $R_t = (\Sigma_{\tau_{\Act}\in \traces_\mathit{fit}}\Sigma_{i=1}^{n_{ep}} R_{t,\tau_{\Act},i}) / (n_{ep} \cdot |\traces_\mathit{fit}|)$
    }
    \SetKwProg{Fn}{Function}{:}{}
    \SetKwFunction{EvalAgent}{EvalAgent}
    \Fn{\EvalAgent{$\pi$, $s_0$, $n_{ep}$}}{
    \For{$i\gets0$ \text{\normalfont\bf\ to } $n_{ep}$}{
    $\tau_i \gets exec_{\pi}(\pi, s_0) = \langle s_0, a_1, r_1 \dots s_n\rangle$ with $s_n\in \states[T]$\;
    $R_{a,i}\gets\Sigma^{n}_{k=1} r_k$ with $r_k \in \tau_i$\;
    }
    \Return $R_a = (\Sigma_{i=1}^{n_{ep}} R_{a,i}) / n_{ep}$
    }
\end{algorithm}

\section{Step 4 - Testing for Performance}
\label{sec:performance}

In the final step, 
we evaluate the performance of trained RL agents. 
The evaluation
compares the accumulated reward 
gained from applying the agent's policy with the accumulated reward gained by executing fuzzed traces. 
Especially in RL settings where the maximal expected reward is unknown, the rewards
gained by the fuzz traces serve as a benchmark for the agents'
performance. 
Furthermore, the fuzz traces are used to test the agents' performance from a diverse set of states.



\begin{algorithm}[t!]
\algorithmfontsize
    \caption{Robust Performance Testing}\label{alg:robust_performance_testing}
    \SetKwInOut{Input}{input}\SetKwInOut{Output}{output}\SetNoFillComment

    \Input{$\MdpInit$, policy $\pi$, fuzz 
    traces $\traces_\mathit{fit}$, \# tests $n_{test}$, 
    \#~episodes $n_{ep}$,
    step width $w$}
    \Output{Avg. accumulated rewards $R_{t}^{pl}$ and $R_{a}^{pl}$} 
    $\textit{pl} \gets w$\;
    \Repeat{$|\{\tau_\Act \in \traces_\mathit{fit}:|\tau_\Act| \geq \textit{pl}\}| < n_{test}$ \tcp*[h]{too few traces of length $pl$}}{ 
    \For{$i \gets 1 \text{\normalfont\bf\ to } n_{test}$}{
    $\tau_\Act \gets \text{ random action trace } \in \traces_\mathit{fit}$\;
    $\tau^{-pl} \gets exec_{\trace}(\tau_{\Act}^{-pl}, s_0) = \langle s_0 \dots s_{pl}\rangle$\;
     $R^- \gets \Sigma^{pl}_{t=1} r_t$ with $r_t \in \tau^{-pl}$\;
    
    $R_{t,i}^{pl} \gets R^- + $\texttt{EvalTraces}($\{\tau^{pl+}_{\Act}\}$, $s_{pl}$, $n_{ep}$)\;  
    
    $R_{a,i}^{pl} \gets R^- + $\texttt{EvalAgent}($\pi$, $ s_{pl}$, $n_{ep}$)\;
    }
    $R_{t}^{pl} \gets (\Sigma_{i=1}^{n_{test}} R_{t,i}^{pl}) / n_{test}$\;
    $R_{a}^{pl} \gets (\Sigma_{i=1}^{n_{test}} R_{a,i}^{pl}) / n_{test}$\;
    
    $\textit{pl} \gets \textit{pl} + w$\;
    } 
    \Return $\bigcup_{pl} \{pl \mapsto R_{t}^{pl}\},\bigcup_{pl} \{pl \mapsto R_{a}^{pl}\}$
    

    
\end{algorithm}

\noindent\textbf{Performance Testing.}
Simple performance testing starts in a fixed initial state and compares the average accumulated reward of the agent with the average accumulated reward resulting from the execution of fuzz traces.
Given the policy $\pi$ of an agent under test, the fuzz traces $\traces_\mathit{fit}$, an initial state $s_0$, and a number of episodes $n_{ep}$, Algo.~\ref{alg:simple_performance_testing} returns the averaged accumulated rewards
of the agent $R_a$ and the fuzz traces $R_t$. 

\noindent\textbf{Robust Performance Testing.}
Robust performance testing targets checking the robustness of learned policies
in potentially unknown situations.
For this purpose, we use the fuzz traces to bring the agent
into a diverse set of states and apply the policy of the agent from these states onward until a terminal state is reached.
To cover states close to the initial state as well as close to the goal, 
we actually use fuzz trace prefixes of increasing length.
The averaged accumulated rewards of the agent traces and the fuzz traces 
serve as performance metric. 

Let $\pi$ be the policy of the RL agent under test, 
$\traces_\mathit{fit}$ be the fuzz traces, 
$s_0$ be an initial state, $w$ be a step width 
to increase the fuzz trace prefix-length, and $n_{test}$ and $n_{ep}$ be numbers of 
tests and episodes, Algo.~\ref{alg:robust_performance_testing} implements our robust performance testing approach.
Starting from prefix length $pl=w$, for each $pl$ the amount of executed tests is $n_{test}$.
To do so, we first select 
a random trace (line 4) and execute its prefix of length 
$pl$ to arrive at a state $s_{pl}$. From $s_{pl}$, we compute 
the accumulated reward of the fuzz trace (line 6) and of the 
agent averaged over $n_{ep}$ episodes (line 8) and add 
the accumulated reward of the common prefix $R^-$ to both.
We average the accumulated rewards over all tests for each
$pl$ individiually (lines 9 and 10) and finally return all results in Line 13.

\pgfplotsset{every tick label/.append style={font=\scriptsize},
label style={font=\scriptsize},
tick label style={font=\scriptsize},
legend style={font=\scriptsize},
legend image code/.code={
\draw[mark repeat=2,mark phase=2]
plot coordinates {
(0cm,0cm)
(0.15cm,0cm)        
(0.3cm,0cm)         
};%
}
}

\section{Experimental Evaluation}
\label{sec:experiments}
We evaluate our testing framework on trained deep RL agents 
for the video game Super Mario Bros., trained 
with varying numbers of training episodes and different
action sets. 

\noindent
\textbf{Setup for RL.} 
We use a third-party
RL agent that operates in the OpenAI gym environment and 
uses double deep Q-Networks~\cite{deep_rl_agent}. Details on the learning parameters along with more experiments and the source code are included in the technical appendix.
To evaluate different agents, we test agents at 
varying stages of learning having three different action sets: 
(1) {\em 2-moves:} fast running and jumping to the right. (2)
{\em basic:}  2-moves plus slow running to the right and left.
(3) {\em right-complex:} basic without running left, but actions for pressing up and down, resulting in the largest action set.
Unless otherwise noted, we present results from training for 
$80k$ episodes. \response{We stopped training at this point since we observed only little improvement from $40k$ to $80k$ episodes, which is also twice as long as suggested~\cite{deep_rl_agent}.}

\noindent
\textbf{Setup for Search and Fuzzing.}
The search for the reference traces uses 
the \emph{2-moves} actions, \response{which are sufficient to
complete most levels in Super Mario Bros. } 
We compute fuzz traces for each action set of the 
different RL agents. We fuzzed for $50$ generations with a population of $50$, used a 
mutation stop-probability $p_{mstop} = 0.2$ with
effect size $ms = 15$ and fitness weights
$\lambda_{cov} = 2$, $\lambda_{pos} = 1.5$, and $\lambda_{neg} = 1$, to focus on exploration.
With a crossover probability of $0.25$, we mostly rely 
on mutations. The search and fuzzing were performed in 
a standard laptop in a few minutes and a few hours, respectively. Compared
to the training that took several days on a dedicated cluster, the computational
effort for testing is relatively low. 
For safety testing, we use $10$ repetitions and a test length of $l=40$
and for performance testing we use $n_{test} = n_{ep} = 10$ and step width $w = 20$.




\noindent

\noindent

\begin{figure}[t]
	\addtocounter{subfigure}{2}
	\centering
	\subfloat[Safety Testing: Relative frequency of fail verdicts]{
\begin{tikzpicture}

\begin{axis}[
tick align=outside,
tick pos=left,
x grid style={white!69.0196078431373!black},
xmin=-0.5, xmax=2.5,
xtick style={color=black},
xtick={0.0,1,2},
xticklabels={
\shortstack{\footnotesize{2-moves} \\ $\:$ \\ $\:$\\ \tiny{$\:$}\\ \tiny{$\:$}},
  \footnotesize{right-complex},
  \footnotesize{basic}
},
height=.22\textheight,
width=.48\textwidth,
ybar=2*\pgflinewidth,
bar width=0.2,
legend columns=2,
y grid style={white!69.0196078431373!black},
ylabel={Fail Verdict Frequency},
ymin=0, ymax=0.65,
ytick style={color=black},
legend style={fill opacity=0.8, draw opacity=1, text opacity=1, draw=white!80!black}
]
\addplot[fill=simplecolor!50]
coordinates
{(0,0.1) (1,0.384615384615385) (2,0.238461538461538)};
\addplot[fill=envcolor!50]
coordinates
{(0,0.0717948717948718) (1,0.351282051282051) (2,0.228205128205128)};
\addplot[fill=comb1color!50]
coordinates
{(0,0.0538461538461538) (1,0.335897435897436) (2,0.23956043956044)};
\addplot[fill=comb2color!50]
coordinates
{(0,0.191025641025641) (1,0.264423076923077) (2,0.33610199324485)};
\legend{simple,  interval ($is=1$), action coverage ($k=1$),
action coverage ($k=2$)}
\end{axis}

\end{tikzpicture}\label{fig:safety_testing}}\\
	\vspace{0.3cm}
	\subfloat[Safety Testing of the \emph{right-complex} agent: Relative frequencies of fail verdicts at boundary states]{
\begin{tikzpicture}

\begin{axis}[
legend cell align={left},
legend style={fill opacity=0.8, draw opacity=1, text opacity=1, draw=white!80!black},
legend style={at={(0.02,0.98)},anchor=north west},
tick align=outside,
tick pos=left,
x grid style={white!69.0196078431373!black},
xmin=-0.6, xmax=12.6,
xtick style={color=black},
y grid style={white!69.0196078431373!black},
ymin=-0.1, ymax=1.3,
height=.2\textheight,
width=.48\textwidth,
xlabel={Boundary State},
ylabel={Fail Verdict Frequency},
ytick style={color=black}
]
\addplot [semithick, simplecolor]
table {%
0 0
1 0
2 0
3 0
4 0
5 1
6 0.9
7 0.1
8 1
9 1
10 0
11 0
12 1
};
\addlegendentry{simple}
\addplot [semithick, envcolor]
table {%
0 0
1 0
2 0
3 0.366666666666667
4 0.333333333333333
5 0.7
6 0.9
7 0
8 0.933333333333333
9 0.666666666666667
10 0
11 0
12 0.666666666666667
};
\addlegendentry{interval ($is=1$)}
\addplot [semithick, comb1color]
table {%
0 0
1 0.3
2 0.0125
3 0.5
4 0.3625
5 0.275
6 0.5875
7 0.0375
8 0.625
9 0.9875
10 0
11 0.0666666666666667
12 0.6125
};
\addlegendentry{action coverage ($k=1$)}
\addplot [semithick, comb2color]
table {%
0 0
1 0.23125
2 0.015625
3 0.246875
4 0.1625
5 0.428125
6 0.578125
7 0.221875
8 0.403125
9 0.528125
10 0.2375
11 0
12 0.384375
};
\addlegendentry{action coverage ($k=2$)}
\end{axis}

\end{tikzpicture}\label{fig:safety_testing_detailed_right_complex}}
	\vspace{-0.2cm}
\end{figure}

\begin{figure}[t]
	\addtocounter{subfigure}{4}
	\centering
	\subfloat[Robust Performance Testing: Average accumulated rewards of fuzz traces and
	the agents trained for $20k$ and $80k$ episodes.]{
\begin{tikzpicture}

\definecolor{color0}{rgb}{0.12156862745098,0.466666666666667,0.705882352941177}
\definecolor{color1}{rgb}{1,0.498039215686275,0.0549019607843137}
\definecolor{color2}{rgb}{0.172549019607843,0.627450980392157,0.172549019607843}
\definecolor{color3}{rgb}{0.83921568627451,0.152941176470588,0.156862745098039}
\definecolor{color4}{rgb}{0.580392156862745,0.403921568627451,0.741176470588235}
\definecolor{color5}{rgb}{0.549019607843137,0.337254901960784,0.294117647058824}
\definecolor{color6}{rgb}{0.890196078431372,0.466666666666667,0.76078431372549}

\begin{axis}[
legend cell align={left},
legend style={fill opacity=0.5, draw opacity=1, text opacity=1, draw=white!80!black},
legend columns=3,
tick align=outside,
tick pos=left,
x grid style={white!69.0196078431373!black},
xmin=2, xmax=380,
ylabel={Average Return},
xtick style={color=black},
xtick={0,40,80,120,160,200,240,280,320,360},
ytick={1000,2000,3000},
xlabel={Fuzz Trace Prefix Length $pl$},
height=.22\textheight,
width=.47\textwidth,
y grid style={white!69.0196078431373!black},
ymin=700, ymax=4000,
ytick style={color=black}
]
\addplot [semithick, twomovescolor,dashed]
table {%
0 1092
20 1190.08
40 1526.8
60 1623.82
80 1723.33
100 1648.95
120 1775.7
140 1560.71
160 1863.48
180 1802.56
200 2136.01
220 2095.58
240 2159.78
260 2205.9
280 2460.88
300 2553.26
320 2759.64
340 2826.34
360 2886.89
380 2932.55
};
\addlegendentry{2-moves ($20k$)}
\addplot [semithick, twomovescolor]
table {%
0 2703
20 2635.54
40 1796.57
60 2010.69
80 2640.74
100 2702.71
120 2798.62
140 2630.77
160 2427.96
180 2423.77
200 2363.24
220 2291.7
240 2493.68
260 2484.71
280 2572.77
300 2641.46
320 2928.65
340 2857.28
360 2938.83
380 2932.12
};
\addlegendentry{2-moves ($80k$)}
\addplot [semithick, complexcolor,dashed]
table {%
0 1568
20 1582.14
40 1513.45
60 869.3
80 1200.21
100 1511.44
120 1657.86
140 1684.68
160 1706.81
180 2031.08
200 2367.74
220 2291.56
240 2402.88
260 2438.94
280 2617.66
300 2709.93
320 2633.88
340 2925.86
360 2905.67
380 2908.09
};
\addlegendentry{right-comp ($20k$)}
\addplot [semithick, complexcolor]
table {%
0 1312
20 1082.2
40 1365.89
60 1530.02
80 2310.54
100 1702.51
120 1809.42
140 2599.63
160 2292.1
180 2465.43
200 2552.9
220 2733.08
240 2691.94
260 2650.78
280 2806.15
300 2838.39
320 2754.96
340 2878.47
360 2899.11
380 2905.71
};
\addlegendentry{right-comp ($80k$)}
\addplot [semithick, basiccolor,dashed]
table {%
0 931
20 962.16
40 879.85
60 821.4
80 877.98
100 1357.45
120 1421.37
140 1627.6
160 1734.88
180 1920.31
200 2030.82
220 2053.2
240 1911.66
260 2099.46
280 2387.21
300 2394.39
320 2436.37
340 2475.61
360 2346.82
380 2609.85
};
\addlegendentry{basic ($20k$)}
\addplot [semithick, basiccolor]
table {%
0 1267
20 1303.45
40 1306.84
60 949.97
80 1211.29
100 1405.35
120 1546.99
140 1527.36
160 1673.85
180 2047.51
200 2128.07
220 2127.58
240 2352.72
260 2426.97
280 2500.57
300 2364.46
320 2448.61
340 2531.03
360 2253.53
380 2647.01
};
\addlegendentry{basic ($80k$)}
\addplot [very thick, fuzzcolor]
table {%
0 2399.81045752 
20 2367.93333333333
40 2402.8
60 2361.68333333333
80 2451.88333333333
100 2376.76666666667
120 2369.63333333333
140 2388.36666666667
160 2475.63333333333
180 2502.18333333333
200 2656.3
220 2655.01666666667
240 2678.83333333333
260 2611.61666666667
280 2759.03333333333
300 2663.66666666667
320 2702.38333333333
340 2846.31666666667
360 2774.11666666667
380 2857.78333333333
};
\addlegendentry{fuzz traces}
\end{axis}

\end{tikzpicture} \label{fig:robust_performance_testing}}
	\vspace{-0.2cm}
	\subfloat[Average frequency of safety violations and average accumulated rewards during testing with the simple test suite.]{\begin{tikzpicture}

\begin{axis}[
tick align=outside,
tick pos=left,
x grid style={white!69.0196078431373!black},
xmin=-0.4, xmax=5.8,
xtick style={color=black},
xtick={0.2,1.2,2.2,3.2,4.2,5.2},
xticklabels={
  \shortstack{2-moves \\ ($20k$)},
  \shortstack{right-c\\ ($20k$)},
  \shortstack{basic\\ ($20k$)},
  \shortstack{2-moves\\ ($80k$)},
  \shortstack{right-c\\ ($80k$)},
  \shortstack{basic\\ ($80k$)}
},
ybar=2*\pgflinewidth,
bar width=0.4,
y grid style={white!69.0196078431373!black},
ylabel={Relative Fail Verdict Frequency},
height=.2\textheight,
width=.41\textwidth,
ymin=0, ymax=0.9,
ytick style={color=black},
legend style={at={(0.025,0.98)},anchor=north west},
legend style={fill opacity=0.8, draw opacity=1, text opacity=1, draw=white!80!black}
]

\addplot[fill=simplecolor!50]
coordinates
{(0,0.523076923076923) (1,0.6) (2,0.315384615384615) 
(3,0.1) (4,0.384615384615385) (5,0.238461538461538)};
\addlegendentry{Safety Violations};

\end{axis}

\begin{axis}[
tick align=outside,
axis y line=right,
ylabel style = {align=center},
x grid style={white!69.0196078431373!black},
xmin=-0.4, xmax=5.8,
xtick style={color=black},
height=.2\textheight,
width=.41\textwidth,
ybar=2*\pgflinewidth,
bar width=0.4,
axis x line=none,
ylabel={\scriptsize Average Return},
ymin=0, ymax=410,
ytick style={color=black},
legend style={at={(0.45,0.98)},anchor=north west},
legend style={fill opacity=0.8, draw opacity=1, text opacity=1, draw=white!80!black}
]

\addplot[fill=envcolor!50]
coordinates
{(0.4,281.319678274428) (1.4,200.579826923077) (2.4,144.377446207019) 
(3.4,406.568123626374) (4.4,270.759519230769) (5.4,204.778416531189)};
\addlegendentry{Average Return};

\end{axis}

\end{tikzpicture}\label{fig:safety_vs_reward}}\\
	\vspace{-0.2cm}
\end{figure}

\textbf{Safety Testing.}
Fig.~\subref{fig:safety_testing} shows the relative number 
of fail verdicts averaged over all tests at all boundary states for agents with different action sets.
For any agent, all test suites found safety violations. 
For instance, the simple test suite produces fail 
verdicts in about $38\%$ of the cases 
when testing the \emph{right-complex} RL agent.
The agent with 2-moves is tested to be the safest agent, failing only $10\%$ of test cases
of the simple test suite.
For  \emph{right-complex}, the least safe agent,
Fig.~\subref{fig:safety_testing_detailed_right_complex} 
depicts
the relative number of fail verdicts distributed over the boundary states, when 
executing all test suites. Note that the results are affected by stochasticity
and they are normalized, so that all results are within $[0,1]$ even though the 
extended test suites perform more tests.
We can see that the early boundary states that are explored 
the most cause the least issues.
Furthermore, we observe that the boundary interval
test suite 
finds safety violations not detected by the simple test suite, e.g.,
at the boundary states $3$ and $4$. 


\noindent
\textbf{Robust Performance Testing.}
We perform robust performance testing 
on all agents trained for $20k$ and $80k$ episodes, respectively.
Fig.~\subref{fig:robust_performance_testing} shows the average 
accumulated rewards (y-axis)
gained by the agents and the fuzz traces when performing fuzz trace
prefixes of the length given by the x-axis. 
It can be seen 
that initially only the well-trained \emph{2-moves} agent surpasses the performance
benchmark set by the fuzz traces. 
Training for $20k$ episodes is not enough for any agent to
achieve rewards close to the fuzz traces and the 
\emph{basic} agent does not improve much with more 
training.
Hence, a larger action space may hurt robustness. The ability 
to move left of the \emph{basic} agent also increases the state space of the underlying MDP, since only moving right induces a DAG-like structure. 
This explains the poor performance of the \emph{basic} agents. 

\noindent\textbf{Relationship between Safety and Performance.}
Finally, we investigate whether performance expressed via 
rewards implies safety. 
Fig.~\subref{fig:safety_vs_reward} shows the average 
number of safety violations 
and the average accumulated rewards gained during testing with 
a simple test suite with all
agents trained for $20k$ and $80k$ episodes, respectively.
Comparing the agents with low amount of training,
the safest agent (\emph{basic}) 
is also the one that gains the lowest reward. 
For well-trained agents, 
the safest agent (\emph{2-moves}) also receives the most reward.
The large negative reward assigned to safety violations
(losing a life) may not be sufficient to enforce safe behavior 
of \emph{right-complex}.
Hence, our testing method may point to issues in reward function design.
However, computing the Pearson correlation coefficient between fail 
verdict frequency and mean accumulated reward for all agents at 
four stages of training with all test suites reveals a moderate negative correlation of $-0.7$, thus high reward often implies low fail frequency.

\section{Concluding Remarks}
\label{sec:conclusion}

We present a search-based testing framework for 
safety and robust-performance testing of RL agents. For safety
testing, we apply backtracking-based DFS to identify 
relevant states and adapt test-adequacy criteria from boundary value
and combinatorial testing. For performance testing, we apply 
genetic-algorithm-based fuzzing starting from a seed trace found by the DFS.
We show both testing methodologies on an off-the-shelf deep RL 
agent for playing Super Mario Bros, where we find safety violations of 
well-trained agents and analyze their performance and robustness. To the best
of our knowledge, we propose one of the first testing frameworks tailored
toward RL. For future work, we will instantiate our framework for 
more RL tasks, where solutions can be found through search and 
other domain-specific approaches. \response{Furthermore, we plan to investigate different fuzzing approaches like fuzzing on the policy level rather than on the trace level.}

\paragraph{Acknowledgments.}
This work has been supported by the "University SAL Labs" initiative of Silicon Austria Labs (SAL) and its Austrian partner universities for applied fundamental research for electronic based systems.
We would like to acknowledge the use of HPC resources
provided by the ZID of Graz University of Technology.
Additionally, this project has received funding from the European Union’s Horizon 2020 research
and innovation programme under grant agreement N$^\circ$ 956123 - FOCETA.
We also thank Vedad Had\v{z}i\'c for his help in the initial development of the 
depth-first search of the reference trace.

\bibliographystyle{named}
\bibliography{ijcai22.bib}

\clearpage
\appendix

\section{Configuration for Deep RL}
The RL agents that we test are based on a PyTorch tutorial~\cite{deep_rl_agent}
and use Double Q-learning to play Super Mario Bros. 
The Deep Q networks consist of a sequence of three convolutional+ReLU 
layers and two fully connected linear layers. 
The input for the first convolutional layer is a 
stack of $4$ transformed images showing the video game screen in $4$ consecutive frames.
The transformation consists of a grey-scale conversion and downsampling from a 
image of size $240$ by $256$ to a square image with $84 \times 84$ pixels.
Stacking $4$ such images enables the agent to detect motion, since it 
is not possible to, e.g.,
detect whether Mario is jumping or landing based
on a single image. 
The output of the final layer is the Q value of an 
action in a state consisting of $4$ stacked,
transformed images.

In the regular, non-sparse reward scheme (see below for results on 
sparse rewards), 
the agent receives positive reward
proportional to the distance travelled in positive horizontal direction (up to $+5$). 
The agent receives negative reward for the time spent in the level. Additionally, it receives a
large constant negative reward for dying, which is $5$ times as large as
the maximal positive reward of a single state-action pair (i.e., $-25$).

The parameters for learning are set as follows for all training runs:
\begin{compactitem}
    \item {\em Discount factor: } $\gamma = 0.9$.
    \item {\em Exploration rate decay: } $\varepsilon_{\mathrm{decay}} = 0.9999995$, the 
    multiplicative factor with which the exploration rate $\varepsilon$ decays
    in the $\varepsilon$-greedy exploration.
    \item {\em Minimum exploration rate: } $\varepsilon_{min} = 0.1$, the smallest
    exploration rate during training.
    \item {\em Burn-in: } $burn = 100$, the number of experiences before training starts.
    \item {\em Batch size: } $batch = 20$, the number of experiences sampled from 
    memory at once for learning.
    \item {\em Memory size: } $mem = 30000$, specifies that up to $mem$ experiences
    are stored in memory, 
    from which we sample for learning.
    \item {\em Learn Interval: } $l_i=3$, the interval between updates of the 
    online Q-network in number of experiences.
    \item {\em Synchronization Interval: } $s_i=1000$,
    the interval between synchronization
    of the online Q-network and the target Q-network in number of experiences.
\end{compactitem}

We fine-tuned the parameters until the learning performance was satisfactory,
and then maintained the parameters unchanged for the training of different agents.
The difference between different trained agents are in action set (2 moves, basic, right-complex), 
training episodes (20k, 40k and 80k) and reward scheme (sparse or dense).

\section{Additional Experimental Results}

\subsection{Testing Agents Trained with Sparse Rewards}
In addition to the agents evaluated in Sec.~\ref{sec:experiments}, we evaluated agents trained using  a sparse reward function.
The \emph{sparse} reward scheme provides reward less often for progressing
in the level. More concretely, it provides rewards when completing a segment of the level ($+10$ at each of five segments),
for defeating enemies ($+1$), and for getting power-ups ($+10)$. 
Since the maximum reward under this scheme is lower,
negative reward from losing lives ($-25$)
should have a larger impact. 
The intuition being that trained agents have a larger 
incentive for safe behavior (gaining power-ups) and avoiding unsafe behavior.

\noindent
\textbf{Safety testing with Sparse Rewards.}
Unfortunately, we did not observe the desired results. 
Fig.~\subref{fig:safety_testing_sparse_20k} shows the relative frequency of fail verdicts
observed during testing after training for $20k$ episodes. 
Fig.~\subref{fig:safety_testing_20k} shows the same for the normal reward scheme after 
training for $20k$ episodes 
(as opposed to $80k$ reported in 
Fig.~\subref{fig:safety_testing}) as a reference. Except in one case, 
testing the \emph{right-complex} with the simple test suite, the sparse rewards lead
to worse behavior w.r.t. safety, despite the fact that the punishment for unsafe
behavior is larger.

\begin{figure}[h]
	\addtocounter{subfigure}{6}
	\centering
	\subfloat[Safety Testing: Relative frequency of fail verdicts under the \emph{sparse}
    reward scheme after training for $20k$ episodes.]{\begin{tikzpicture}

\begin{axis}[
tick align=outside,
tick pos=left,
x grid style={white!69.0196078431373!black},
xmin=-0.5, xmax=2.5,
xtick style={color=black},
xtick={0.0,1,2},
xticklabels={
  2-moves,
  right-complex,
  basic
},
height=.24\textheight,
width=.48\textwidth,
ybar=2*\pgflinewidth,
bar width=0.2,
legend columns=2,
y grid style={white!69.0196078431373!black},
ylabel={Fail Verdict Frequency},
ymin=0, ymax=1,
ytick style={color=black},
legend style={fill opacity=0.8, draw opacity=1, text opacity=1, draw=white!80!black}
]
\addplot[fill=simplecolor!50]
coordinates
{(0,0.776923076923077) (1,0.553846153846154) (2,0.6)};
\addplot[fill=envcolor!50]
coordinates
{(0,0.692307692307692) (1,0.561538461538462) (2,0.528205128205128)};
\addplot[fill=comb1color!50]
coordinates
{(0,0.653846153846154) (1,0.563141025641026) (2,0.531868131868132)};
\addplot[fill=comb2color!50]
coordinates
{(0,0.630769230769231) (1,0.630306952662722) (2,0.537548166119595)};
\legend{simple,  environment ($e=1$), action coverage ($k=1$),
action coverage ($k=2$)}
\end{axis}

\end{tikzpicture}\label{fig:safety_testing_sparse_20k}}
    
	\subfloat[Safety Testing: Relative frequency of fail verdicts under the normal
    reward scheme after training for $20k$ episodes.]{\begin{tikzpicture}

\begin{axis}[
tick align=outside,
tick pos=left,
x grid style={white!69.0196078431373!black},
xmin=-0.5, xmax=2.5,
xtick style={color=black},
xtick={0.0,1,2},
xticklabels={
  2-moves,
  right-complex,
  basic
},
height=.24\textheight,
width=.48\textwidth,
ybar=2*\pgflinewidth,
bar width=0.2,
legend columns=2,
y grid style={white!69.0196078431373!black},
ylabel={Fail Verdict Frequency},
ymin=0, ymax=1,
ytick style={color=black},
legend style={fill opacity=0.8, draw opacity=1, text opacity=1, draw=white!80!black}
]
\addplot[fill=simplecolor!50]
coordinates
{(0,0.523076923076923) (1,0.6) (2,0.315384615384615)};
\addplot[fill=envcolor!50]
coordinates
{(0,0.464102564102564) (1,0.543589743589744) (2,0.325641025641026)};
\addplot[fill=comb1color!50]
coordinates
{(0,0.457692307692308) (1,0.444230769230769) (2,0.368131868131868)};
\addplot[fill=comb2color!50]
coordinates
{(0,0.435897435897436) (1,0.391327662721894) (2,0.374473145901717)};
\legend{simple,  interval ($is=1$), action coverage ($k=1$),
action coverage ($k=2$)}
\end{axis}

\end{tikzpicture}\label{fig:safety_testing_20k}}
\end{figure}

\addtocounter{figure}{6}

\begin{figure}[t]
    \centering
\begin{tikzpicture}

\definecolor{color0}{rgb}{0.12156862745098,0.466666666666667,0.705882352941177}
\definecolor{color1}{rgb}{1,0.498039215686275,0.0549019607843137}
\definecolor{color2}{rgb}{0.172549019607843,0.627450980392157,0.172549019607843}
\definecolor{color3}{rgb}{0.83921568627451,0.152941176470588,0.156862745098039}

\begin{axis}[
legend cell align={left},
legend style={fill opacity=0.8, draw opacity=1, text opacity=1, draw=white!80!black},
legend style={at={(0.02,0.98)},anchor=north west},
tick align=outside,
tick pos=left,
x grid style={white!69.0196078431373!black},
xmin=-0.6, xmax=12.6,
xtick style={color=black},
y grid style={white!69.0196078431373!black},
ymin=-0.1, ymax=1.6,
height=.22\textheight,
width=.48\textwidth,
xlabel={Boundary State},
ylabel={Fail Verdict Frequency},
ytick style={color=black}
]
\addplot [semithick, simplecolor]
table {%
0 0
1 1
2 1
3 1
4 0
5 1
6 1
7 1
8 0.1
9 1
10 0
11 0.1
12 0
};
\addlegendentry{simple}
\addplot [semithick, envcolor]
table {%
0 0
1 0.933333333333333
2 1
3 1
4 0
5 0.966666666666667
6 1
7 1
8 0.4
9 0.966666666666667
10 0.0333333333333333
11 0
12 0
};
\addlegendentry{interval ($is=1$)}
\addplot [semithick, comb1color]
table {%
0 0
1 0.9625
2 0.975
3 0.75
4 0
5 1
6 0.975
7 0.9375
8 0.725
9 0.9625
10 0
11 0.0333333333333333
12 0
};
\addlegendentry{action coverage ($k=1$)}
\addplot [semithick, comb2color]
table {%
0 0
1 0.603125
2 0.984375
3 0.853125
4 0.26875
5 0.99375
6 1
7 0.971875
8 0.91875
9 0.975
10 0.240625
11 0.384615384615385
12 0
};
\addlegendentry{action coverage ($k=2$)}
\end{axis}

\end{tikzpicture}
    \caption{Safety Testing of the \emph{right-complex} agent with sparse rewards: Relative frequencies of fail verdicts at boundary states.}
    \label{fig:safety_testing_detailed_complex_sparse}
\end{figure}
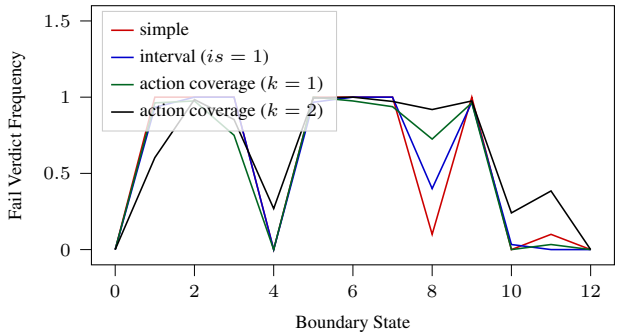

Analyzing the behavior in more detail, 
we see an interesting effect, though. Fig.~\ref{fig:safety_testing_detailed_complex_sparse} shows the fail verdicts
frequencies at the boundary states from safety testing the \emph{right-complex}
agent with sparse reward. While the agent trained with normal rewards (Fig.~\subref{fig:safety_testing_detailed_right_complex}) showed unsafe behavior
at boundary state $12$, training with sparse reward led to safe behavior at
this state. At most other states, especially at the beginning, sparse rewards 
were detrimental to safety. 

Fig.~\ref{fig:safety_testing_detailed_complex_sparse} includes an effect
of stochasticity that we want to briefly explain. At boundary state $11$,
the simple test suite finds a non-zero amount of issues and the environment test suite
does not find any issues, even though the latter includes the former. This can 
be explained by stochastic behavior, causing one execution of the test case
at boundary state $11$ to find an issue that was not detected while executing
the environment test suite. 

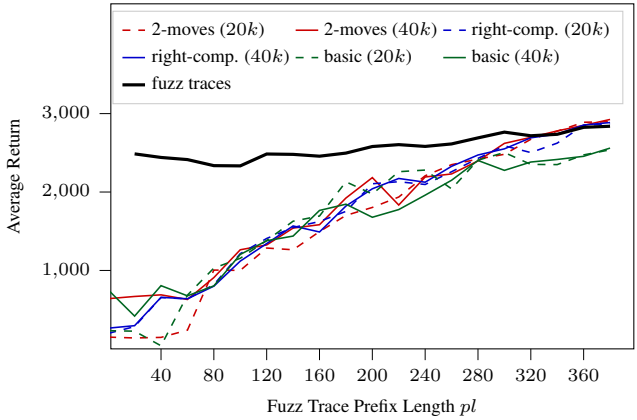
\begin{figure}[t]
    \centering
\begin{tikzpicture}

\definecolor{color0}{rgb}{0.12156862745098,0.466666666666667,0.705882352941177}
\definecolor{color1}{rgb}{1,0.498039215686275,0.0549019607843137}
\definecolor{color2}{rgb}{0.172549019607843,0.627450980392157,0.172549019607843}
\definecolor{color3}{rgb}{0.83921568627451,0.152941176470588,0.156862745098039}
\definecolor{color4}{rgb}{0.580392156862745,0.403921568627451,0.741176470588235}
\definecolor{color5}{rgb}{0.549019607843137,0.337254901960784,0.294117647058824}
\definecolor{color6}{rgb}{0.890196078431372,0.466666666666667,0.76078431372549}

\begin{axis}[
legend cell align={left},
legend style={fill opacity=0.5, draw opacity=1, text opacity=1, draw=white!80!black},
legend columns=3,
tick align=outside,
tick pos=left,
x grid style={white!69.0196078431373!black},
xmin=2, xmax=398,
ylabel={Average Return},
xtick style={color=black},
xtick={0,40,80,120,160,200,240,280,320,360},
ytick={1000,2000,3000},
xlabel={Fuzz Trace Prefix Length $pl$},
height=.27\textheight,
width=.48\textwidth,
y grid style={white!69.0196078431373!black},
ymin=0, ymax=4400,
ytick style={color=black}
]
\addplot [semithick, twomovescolor, dashed]
table {%
0 150
20 139
40 146.27
60 238.92
80 1004.79
100 1000.78
120 1285.81
140 1263.65
160 1495.28
180 1701.9
200 1803.37
220 1936.29
240 2207.15
260 2348.96
280 2420.47
300 2481.19
320 2672.58
340 2775.69
360 2889.63
380 2896.19
};
\addlegendentry{2-moves ($20k$)}
\addplot [semithick,twomovescolor]
table {%
0 640
20 668.27
40 688.87
60 629.7
80 909.01
100 1263.67
120 1325.28
140 1540.32
160 1583.27
180 1923.4
200 2183.97
220 1834.22
240 2191.68
260 2230.59
280 2396.58
300 2620.49
320 2696.72
340 2779.5
360 2850.13
380 2925.33
};
\addlegendentry{2-moves ($40k$)}
\addplot [semithick, complexcolor,dashed]
table {%
0 197
20 284.03
40 655.18
60 636.57
80 818.32
100 1214.18
120 1407.76
140 1547.23
160 1624.31
180 1753.23
200 2105.59
220 2131.43
240 2096.21
260 2258.47
280 2429.23
300 2589.2
320 2503.15
340 2623.73
360 2850.82
380 2869.9
};
\addlegendentry{right-comp. ($20k$)}
\addplot [semithick, complexcolor]
table {%
0 264
20 295.8
40 656.88
60 638.84
80 799.67
100 1118.93
120 1342.68
140 1567.32
160 1490.64
180 1821.14
200 2041.29
220 2172.88
240 2125.26
260 2328.06
280 2472.06
300 2550.73
320 2691.4
340 2738.78
360 2853.79
380 2886.11
};
\addlegendentry{right-comp. ($40k$)}
\addplot [semithick, basiccolor,dashed]
table {%
0 232
20 225.49
40 39.31
60 685.54
80 1029.83
100 1155.04
120 1382.49
140 1627.88
160 1694.18
180 2130.52
200 1974.47
220 2259.22
240 2280.09
260 2040.2
280 2406.61
300 2503.7
320 2352.31
340 2349.77
360 2473.35
380 2538.24
};
\addlegendentry{basic ($20k$)}
\addplot [semithick, basiccolor]
table {%
0 752
20 417.77
40 805.98
60 676.32
80 803.18
100 1203.83
120 1378.73
140 1438.3
160 1765.13
180 1844.86
200 1678.24
220 1776.29
240 1960.08
260 2150.21
280 2402.36
300 2276.24
320 2382.57
340 2415.34
360 2455.56
380 2560.15
};
\addlegendentry{basic ($40k$)}
\addplot [very thick, fuzzcolor]
table {%
20 2486.48333333333
40 2441
60 2413.65
80 2336.45
100 2333.43333333333
120 2484.86666666667
140 2480.21666666667
160 2457
180 2496.5
200 2580.36666666667
220 2604.4
240 2581.73333333333
260 2613.35
280 2690.51666666667
300 2763.8
320 2717.56666666667
340 2736.18333333333
360 2824.91666666667
380 2838.88333333333
};
\addlegendentry{fuzz traces}
\end{axis}

\end{tikzpicture}
    \caption{Performance testing of agents trained with sparse rewards.}
    \label{fig:performance_sparse}
\end{figure}

\noindent
\textbf{Performance testing with Sparse Rewards.}
A possible explanation for unsafe behavior may simply be that training 
with sparse rewards worked less well, resulting in worse-performing agents. 
Fig.~\ref{fig:performance_sparse} shows
the results of performance testing these agents after $20k$ and $40k$ of training,
respectively. In the performance testing, we apply the normal reward scheme to 
enable a better comparison with the results present in Sect.~\ref{sec:experiments}.
In both cases, the agents are trained to solve the same task, thus 
well-trained agents should perform well with either reward scheme.
None of the agents' performance comes close to the reward gained
by the fuzz traces, except for prefix lengths greater than $280$, i.e., very 
late in the Super Mario Bros.\ level. 
Since we observed that the performance difference 
between training for $20k$ and $40k$ is consistently relatively low, 
we refrained from training the agents with sparse rewards for the full
$80k$ episodes used for the normal rewards.

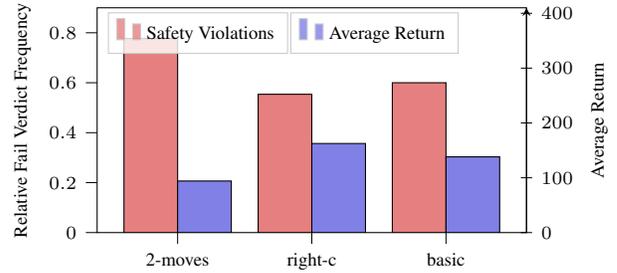
\begin{figure}[t]
    \centering
    \begin{tikzpicture}

\begin{axis}[
tick align=outside,
tick pos=left,
x grid style={white!69.0196078431373!black},
xmin=-0.4, xmax=2.8,
xtick style={color=black},
xtick={0.2,1.2,2.2,3.2,4.2,5.2},
xticklabels={
  2-moves, 
  right-c,
  basic
},
ybar=2*\pgflinewidth,
bar width=0.4,
y grid style={white!69.0196078431373!black},
ylabel={Relative Fail Verdict Frequency},
height=.2\textheight,
width=.41\textwidth,
ymin=0, ymax=0.9,
ytick style={color=black},
legend style={at={(0.025,0.98)},anchor=north west},
legend style={fill opacity=0.8, draw opacity=1, text opacity=1, draw=white!80!black}
]

\addplot[fill=simplecolor!50]
coordinates
{(0,0.776923076923077) (1,0.553846153846154) (2,0.6)};
\addlegendentry{Safety Violations};

\end{axis}

\begin{axis}[
tick align=outside,
axis y line=right,
ylabel style = {align=center},
x grid style={white!69.0196078431373!black},
xmin=-0.4, xmax=2.8,
xtick style={color=black},
height=.2\textheight,
width=.41\textwidth,
ybar=2*\pgflinewidth,
bar width=0.4,
axis x line=none,
ylabel={\scriptsize Average Return},
ymin=0, ymax=410,
ytick style={color=black},
legend style={at={(0.45,0.98)},anchor=north west},
legend style={fill opacity=0.8, draw opacity=1, text opacity=1, draw=white!80!black}
]

\addplot[fill=envcolor!50]
coordinates
{(0.4,94.1051324203092) (1.4,162.516564032122) (2.4,138.176606768232) };
\addlegendentry{Average Return};

\end{axis}

\end{tikzpicture}
    \caption{Average frequency of safety violations (fails verdicts) and average accumulated reward during testing of the sparse-reward agents with the simple test suite.}
    \label{fig:app_safety_vs_reward_sparse}
\end{figure}

\noindent
\textbf{Comparing Safety and Performance.}
Comparing safety and performance during safety testing in 
Fig.~\ref{fig:app_safety_vs_reward_sparse}, it can be seen that 
the safest agents receives the highest reward and 
the second and third safest receive the 
second-highest and third-highest rewards.  

\subsection{Testing Agent Trained for 1-2}
\begin{figure}[t]
    \centering
    \begin{tikzpicture}

\begin{axis}[
tick align=outside,
tick pos=left,
x grid style={white!69.0196078431373!black},
xmin=-0.5, xmax=2.5,
xtick style={color=black},
xtick={0.0,1,2},
xticklabels={
  2-moves,
  right-complex,
  basic
},
height=.24\textheight,
width=.48\textwidth,
ybar=2*\pgflinewidth,
bar width=0.2,
legend columns=2,
y grid style={white!69.0196078431373!black},
ylabel={Fail Verdict Frequency},
ymin=0, ymax=1,
ytick style={color=black},
legend style={fill opacity=0.8, draw opacity=1, text opacity=1, draw=white!80!black}
]
\addplot[fill=simplecolor!50]
coordinates
{(0,0.455555555555555) (1,0.638888888888889) (2,0.705555555555556)};
\addplot[fill=envcolor!50]
coordinates
{(0,0.485185185185185) (1,0.562962962962963) (2,0.72037037037037)};
\addplot[fill=comb1color!50]
coordinates
{(0,0.563888888888889) (1,0.604861111111111) (2,0.698412698412698)};
\addplot[fill=comb2color!50]
coordinates
{(0,0.584722222222222) (1,0.0) (2,0.672675736961451)};
\legend{simple,  interval ($is=1$), action coverage ($k=1$),
action coverage ($k=2$)}
\end{axis}

\end{tikzpicture}
    \caption{Safety Testing: Relative frequency of fail verdicts when testing agents
    trained to complete level 1-2 after training for $80k$ episodes.}
    \label{fig:safety_testing_1_2}
\end{figure}
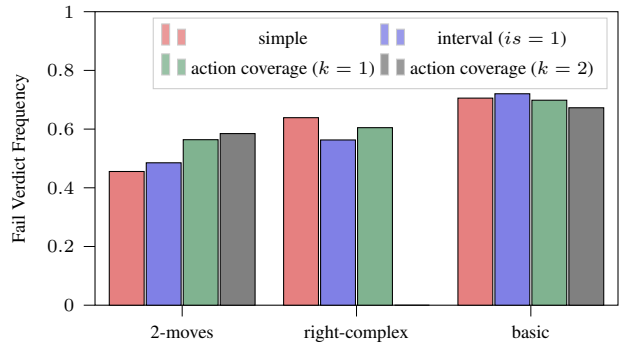
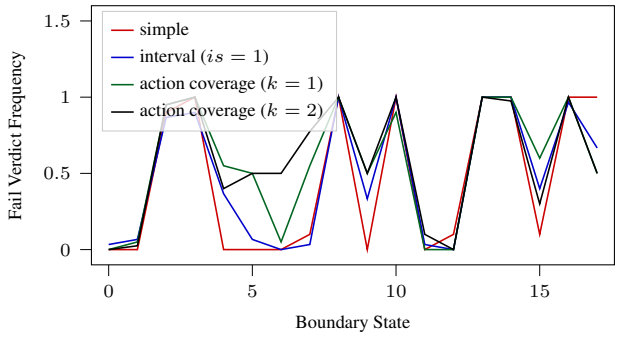
\begin{figure}[t]
    \centering
\begin{tikzpicture}

\definecolor{color0}{rgb}{0.12156862745098,0.466666666666667,0.705882352941177}
\definecolor{color1}{rgb}{1,0.498039215686275,0.0549019607843137}
\definecolor{color2}{rgb}{0.172549019607843,0.627450980392157,0.172549019607843}
\definecolor{color3}{rgb}{0.83921568627451,0.152941176470588,0.156862745098039}

\begin{axis}[
legend cell align={left},
legend style={fill opacity=0.8, draw opacity=1, text opacity=1, draw=white!80!black},
legend style={at={(0.02,0.98)},anchor=north west},
tick align=outside,
tick pos=left,
x grid style={white!69.0196078431373!black},
xmin=-0.6, xmax=17.6,
xtick style={color=black},
y grid style={white!69.0196078431373!black},
ymin=-0.1, ymax=1.6,
height=.22\textheight,
width=.48\textwidth,
xlabel={Boundary State},
ylabel={Fail Verdict Frequency},
ytick style={color=black}
]
\addplot [semithick, simplecolor]
table {%
0 0
1 0
2 0.9
3 1
4 0
5 0
6 0
7 0.1
8 1
9 0
10 1
11 0
12 0.1
13 1
14 1
15 0.1
16 1
17 1
};
\addlegendentry{simple}
\addplot [semithick, envcolor]
table {%
0 0.0333333333333333
1 0.0666666666666667
2 0.866666666666667
3 0.9
4 0.366666666666667
5 0.0666666666666667
6 0
7 0.0333333333333333
8 1
9 0.333333333333333
10 1
11 0.0333333333333333
12 0
13 1
14 1
15 0.4
16 0.966666666666667
17 0.666666666666667
};
\addlegendentry{interval ($is=1$)}
\addplot [semithick, comb1color]
table {%
0 0
1 0.05
2 0.95
3 1
4 0.55
5 0.5
6 0.05
7 0.55
8 1
9 0.5
10 0.9
11 0
12 0
13 1
14 1
15 0.6
16 1
17 0.5
};
\addlegendentry{action coverage ($k=1$)}
\addplot [semithick, comb2color]
table {%
0 0
1 0.025
2 0.95
3 1
4 0.4
5 0.5
6 0.5
7 0.775
8 1
9 0.5
10 1
11 0.1
12 0
13 1
14 0.975
15 0.3
16 1
17 0.5
};
\addlegendentry{action coverage ($k=2$)}
\end{axis}

\end{tikzpicture}
    \caption{Safety Testing of the \emph{2-moves} agent trained to complete level 1-2: Relative frequencies of fail verdicts at boundary states.}
    \label{fig:safety_2moves_detailed_1_2}
\end{figure}
In addition to agents trained to complete the first level 
of Super Mario Bros., we also tested agents trained to complete
the second level, which is called 1-2. For this case study, 
we again used the normal reward scheme. 

\noindent
\textbf{Safety Testing.}
Fig.~\ref{fig:safety_testing_1_2}
shows a summary of the results of safety testing RL agents that have been
trained for $80k$ episodes. The execution of the second action 
coverage test suite
with $k=2$ on the \emph{right-complex} agent unfortunately 
did not finish in time for the submission. 
It can be seen that the agents performed 
considerably worse w.r.t.\ safety than the agents trained for the 
first level. Looking at the detailed results for the \emph{2-moves}
in Fig.~\ref{fig:safety_2moves_detailed_1_2} presents a 
possible explanation. The level contains more boundary states, i.e., 
more safety-critical situations that the agents need to deal with.
There are several states, where the \emph{2-moves} agent, 
despite being the safest,
has a relative fail frequency close to $1$. 

\begin{figure}[t]
    \centering
\begin{tikzpicture}

\definecolor{color0}{rgb}{0.12156862745098,0.466666666666667,0.705882352941177}
\definecolor{color1}{rgb}{1,0.498039215686275,0.0549019607843137}
\definecolor{color2}{rgb}{0.172549019607843,0.627450980392157,0.172549019607843}
\definecolor{color3}{rgb}{0.83921568627451,0.152941176470588,0.156862745098039}
\definecolor{color4}{rgb}{0.580392156862745,0.403921568627451,0.741176470588235}
\definecolor{color5}{rgb}{0.549019607843137,0.337254901960784,0.294117647058824}
\definecolor{color6}{rgb}{0.890196078431372,0.466666666666667,0.76078431372549}

\begin{axis}[
legend cell align={left},
legend style={fill opacity=0.5, draw opacity=1, text opacity=1, draw=white!80!black},
legend columns=3,
tick align=outside,
tick pos=left,
x grid style={white!69.0196078431373!black},
xmin=20, xmax=420,
ylabel={Average Return},
xtick style={color=black},
xtick={0,40,80,120,160,200,240,280,320,360,400},
ytick={1000,2000,3000},
xlabel={Fuzz Trace Prefix Length $pl$},
height=.27\textheight,
width=.48\textwidth,
y grid style={white!69.0196078431373!black},
ymin=0, ymax=3800,
ytick style={color=black}
]
\addplot [semithick, twomovescolor]
table {%
20 1467.23
40 1400.36
60 1528.51
80 1308.2
100 1173.59
120 1446.1
140 1485.87
160 1532.31
180 1844.54
200 2198.29
220 1997.95
240 2314.52
260 2312.94
280 2198.4
300 2268.93
320 2511.42
340 2437.47
360 2509.15
380 2489.27
400 2533.07
420 2569.68
};
\addlegendentry{2-moves ($80k$)}
\addplot [semithick, twomovescolor,dashed]
table {%
20 1739.6
40 1213.63
60 1683.81
80 1586.3
100 1398.37
120 1925.89
140 2386.56
160 2350.75
180 2089.39
200 2002.79
220 2206.83
240 2266.52
260 2201.67
280 2543.07
300 2356.39
320 2313.65
340 2448.42
360 2340.81
380 2587.99
400 2636.77
420 2518.95
};
\addlegendentry{2-moves ($40k$)}
\addplot [semithick, complexcolor]
table {%
20 1003.96
40 537.96
60 913.34
80 935.69
100 1228.49
120 1701.36
140 1496.43
160 1836.15
180 1858.75
200 1855.21
220 1952.83
240 1940.45
260 2193.57
280 2322.95
300 2493.84
320 2539.2
340 2564.96
360 2531.79
380 2505.5
400 2610.79
420 2565.82
};
\addlegendentry{right-comp ($80k$)}
\addplot [semithick, complexcolor, dashed]
table {%
20 1414.37
40 892.32
60 618.98
80 805.26
100 1131.72
120 1743.72
140 1757.73
160 1786.03
180 1916.92
200 1853.34
220 1898.87
240 2002.19
260 2101.15
280 2325.58
300 2448.65
320 2309.43
340 2434.17
360 2432.84
380 2458.7
400 2507.04
420 2451.45
};
\addlegendentry{right-comp ($40k$)}
\addplot [semithick, basiccolor]
table {%
20 1142.6
40 1357.29
60 1026.77
80 990.87
100 981.68
120 1121.53
140 1665.64
160 1807.9
180 1681.9
200 2131.71
220 1878.93
240 1917.27
260 2000.26
280 2152.22
300 1934.79
320 2014.74
340 2068.33
360 2336.72
380 2296.84
400 2276.38
420 2413.47
};
\addlegendentry{basic ($80k$)}
\addplot [semithick, basiccolor,dashed]
table {%
20 1014.23
40 1101.52
60 932.83
80 912.56
100 1528.07
120 1503.26
140 1896.97
160 1851.62
180 2065.25
200 1629.27
220 1930.4
240 1928.49
260 2026.49
280 2235.86
300 1974.92
320 1969.88
340 2127.41
360 2156.79
380 2174.73
400 2299.39
420 2229.08
};
\addlegendentry{basic ($40k$)}
\addplot [very thick, fuzzcolor]
table {%
20 2277.53333333333
40 2086.1
60 2282.45
80 2118.11666666667
100 2186.81666666667
120 2213.5
140 2262.18333333333
160 2339.36666666667
180 2300.7
200 2376.65
220 2315.85
240 2339.93333333333
260 2435.75
280 2550.86666666667
300 2504.28333333333
320 2524.08333333333
340 2547
360 2614.05
380 2589.95
400 2624.7
420 2617.36666666667
};
\addlegendentry{fuzz traces}
\end{axis}

\end{tikzpicture}
    \caption{Performance testing of agents trained to complete level
    1-2.}
    \label{fig:performance_1_2}
\end{figure}
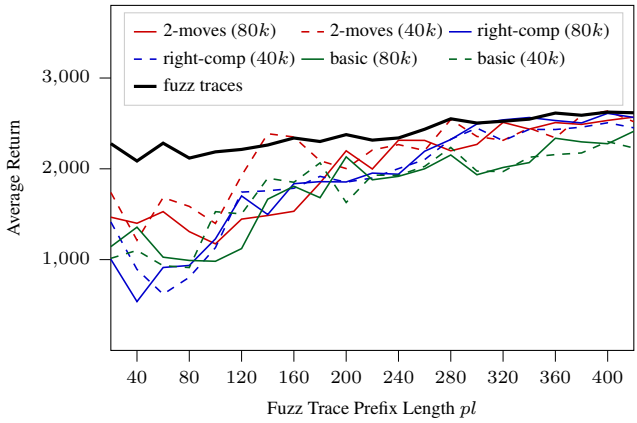
\noindent
\textbf{Performance Testing.} Fig.~\ref{fig:performance_1_2} shows 
the results from robust performance testing. 
The agents do not achieve the performance benchmark set by the fuzz 
testing until very late in the level, similarly to the sparse reward 
scheme. At the stage where the agents' performance results are close to 
the fuzz trace performance, several safety-critical situations have
been already passed by the fuzz trace prefix. In contrast to the 
first level, the \emph{basic} agent performs similarly well as the others
and increased training does not help performance; the \emph{2-moves}
trained for $40k$ episodes even performs better than when trained for 
$80k$ episodes. 

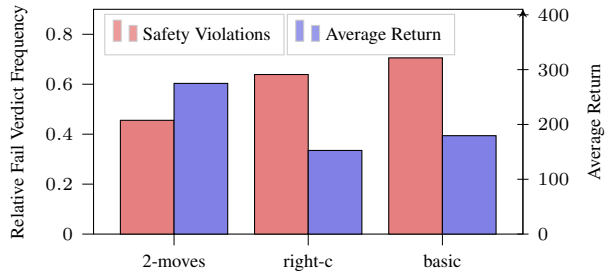
\begin{figure}[t!]
    \centering
    \begin{tikzpicture}

\begin{axis}[
tick align=outside,
tick pos=left,
x grid style={white!69.0196078431373!black},
xmin=-0.4, xmax=2.8,
xtick style={color=black},
xtick={0.2,1.2,2.2,3.2,4.2,5.2},
xticklabels={
  2-moves, 
  right-c,
  basic
},
ybar=2*\pgflinewidth,
bar width=0.4,
y grid style={white!69.0196078431373!black},
ylabel={Relative Fail Verdict Frequency},
height=.2\textheight,
width=.41\textwidth,
ymin=0, ymax=0.9,
ytick style={color=black},
legend style={at={(0.025,0.98)},anchor=north west},
legend style={fill opacity=0.8, draw opacity=1, text opacity=1, draw=white!80!black}
]

\addplot[fill=simplecolor!50]
coordinates
{(0,0.455555555555555) (1,0.638888888888889) (2,0.705555555555556)};
\addlegendentry{Safety Violations};

\end{axis}

\begin{axis}[
tick align=outside,
axis y line=right,
ylabel style = {align=center},
x grid style={white!69.0196078431373!black},
xmin=-0.4, xmax=2.8,
xtick style={color=black},
height=.2\textheight,
width=.41\textwidth,
ybar=2*\pgflinewidth,
bar width=0.4,
axis x line=none,
ylabel={\scriptsize Average Return},
ymin=0, ymax=410,
ytick style={color=black},
legend style={at={(0.45,0.98)},anchor=north west},
legend style={fill opacity=0.8, draw opacity=1, text opacity=1, draw=white!80!black}
]

\addplot[fill=envcolor!50]
coordinates
{(0.4,274.881818961819) (1.4,152.567938948307) (2.4,179.517000891266) };
\addlegendentry{Average Return};

\end{axis}

\end{tikzpicture}
    \caption{Average frequency of safety violations (fails verdicts) and average accumulated reward during testing of agents trained to complete level 1-2 with the simple test suite.}
    \label{fig:app_safety_vs_reward_1_2}
\end{figure}
\noindent
\textbf{Comparing Safety and Performance.}
Finally, we want to compare safety and performance during safety
testing for these agents as well. Fig.~\ref{fig:app_safety_vs_reward_1_2}
shows the average accumulated rewards and the relative fail frequency
during safety testing with the simple test suite. As for the first
level, the \emph{2-moves} agents is the safest and achieves the highest
reward. Apart from that, the difference between the other two agents 
is relatively small. This shows that the small action space enables 
considerably better training.

\end{document}